%% file: main_aaai.tex
\documentclass[10pt,twocolumn,letterpaper]{article}

\usepackage{cvpr}              % To produce the CAMERA-READY version
\usepackage{amsmath}
\usepackage{amssymb}
\usepackage{booktabs}

\usepackage{adjustbox}
\usepackage{algorithm}
\usepackage{algpseudocode}
\usepackage[utf8]{inputenc}
\usepackage{nicefrac}
\usepackage{tabu}
\usepackage{lipsum}
\usepackage{multirow}

\usepackage{arydshln}

\usepackage[pagebackref,breaklinks,colorlinks]{hyperref}
\usepackage[capitalize]{cleveref}
\crefname{section}{Sec.}{Secs.}
\Crefname{section}{Section}{Sections}
\Crefname{table}{Table}{Tables}
\crefname{table}{Tab.}{Tabs.}

\def\cvprPaperID{*****} % *** Enter the CVPR Paper ID here
\def\confName{CVPR}
\def\confYear{2022}

\begin{document}

% \title{\LaTeX\ Author Guidelines for \confName~Proceedings}
\title{Self-supervised GAN Detector}

\author{Yonghyun Jeong$^{1}$, Doyeon Kim$^{2}$, Pyounggeon Kim$^{2}$, Youngmin Ro$^{3}$, Jongwon Choi$^{4}$\\
$^{1}$NAVER Clova, $^{2}$Samsung SDS, $^{3}$University of Seoul, $^{4}$Chung-Ang University\\
{\tt\small  yonghyun.jeong@navercorp.com, dy31.kim@samsung.com, siru.kim@samsung.com,} \\
{\tt\small youngmin.ro@uoa.ac.kr, choijw@cau.ac.kr}
}
\maketitle

% \AddToShipoutPicture*{%
%      \AtTextUpperLeft{%
%          \put(0,30){
%           \begin{minipage}{\textwidth}
%               \footnotesize
%               Preprint version\\
%               Under review of Pattern Recognition Letters (Elsevier)\\
%           \end{minipage}}%
%      }%
% }

\input{0_abstract}

\input{1_intro}

\input{2_related_work}
\input{3_method}
\input{4_result}
\input{5_conclusion}
\small
\bibliographystyle{unsrt}
\bibliography{refs}

\end{document}

%% file: 0_abstract.tex
\begin{abstract}
Although the recent advancement in generative models brings diverse advantages to society, it can also be abused with malicious purposes, such as fraud, defamation, and fake news. To prevent such cases, vigorous research is conducted to distinguish the generated images from the real images, but challenges still remain to distinguish the unseen generated images outside of the training settings. Such limitations occur due to data dependency arising from the model's overfitting issue to the training data generated by specific GANs. To overcome this issue, we adopt a self-supervised scheme to propose a novel framework. Our proposed method is composed of the artificial fingerprint generator reconstructing the high-quality artificial fingerprints of GAN images for detailed analysis, and the GAN detector distinguishing GAN images by learning the reconstructed artificial fingerprints. To improve the generalization of the artificial fingerprint generator, we build multiple autoencoders with different numbers of upconvolution layers. With numerous ablation studies, the robust generalization of our method is validated by outperforming the generalization of the previous state-of-the-art algorithms, even without utilizing the GAN images of the training dataset.
\end{abstract}
%{\let\thefootnote\relax\footnotetext{Under Review (Pattern Recognition Letters)}}

%% file: 1_intro.tex
\begin{figure*}[]
\centering
\includegraphics[width=0.7\linewidth]{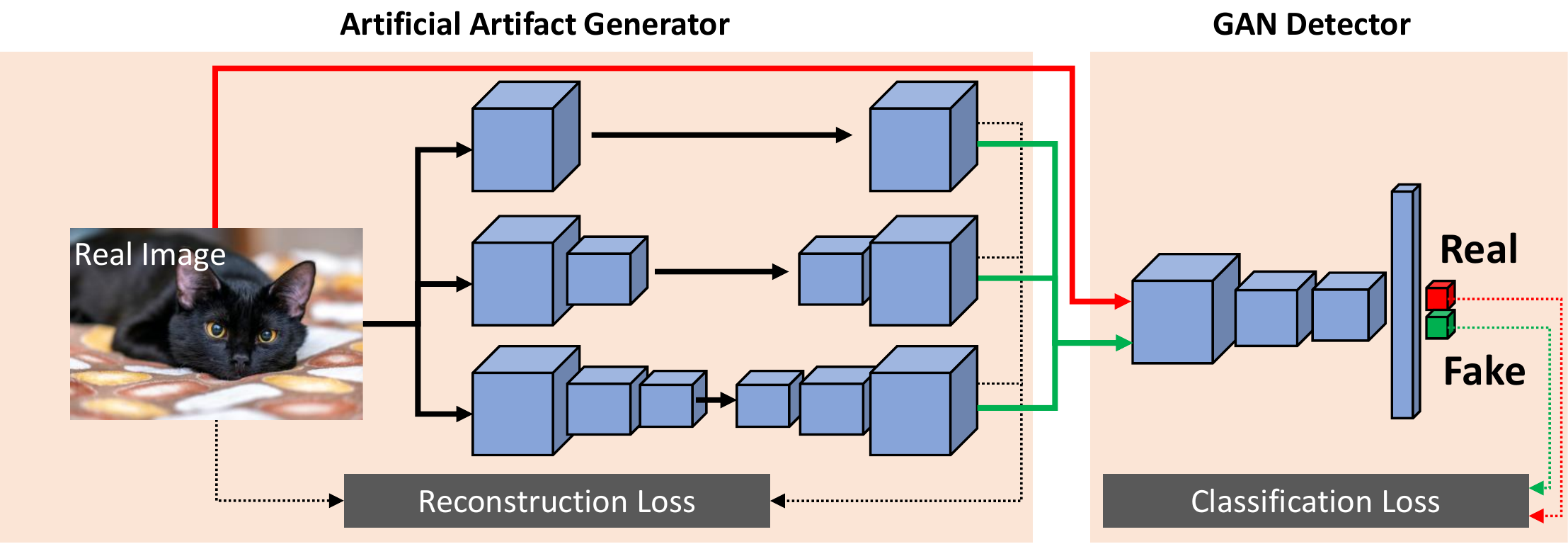}
\caption{\textbf{Overall Framework.}
The proposed framework consists of two modules including the artificial fingerprint generator and the GAN detector. The artificial fingerprint generator contains multiple autoencoders trained by the reconstruction loss, while the GAN detector is a conventional classification network to distinguish the real images from the reconstructed images from the artificial fingerprint generator.}
\label{fig:framework}
%\vspace{-1.5em}
\end{figure*}

\section{Introduction}\label{intro}
Based on the recent enhancement of the generative models, such as Generative Adversarial Networks (GAN)~\cite{gan}, it has become easy to obtain high-quality fake images generated by deep-learning models~\cite{stylegan,stylegan2}. 
Many recent generative models can even transform the target images to include the specific properties of the users' choices~\cite{cyclegan,stargan,stargan2,alae,id_invert}.
However, as technology progresses, the danger of intentionally misusing the fake images for purposes like fraud, defamation, and fake news grows~\cite{lee,kwon,nguyen2019deep}.
To prevent such cases, it is important to distinguish between the real images and the generated images by GANs~\cite{sun}.

Previous image forgery detection techniques tended to focus on unnatural features in human faces~\cite{headpose,facexray}, but many recent models have evolved to uncover distinctive features produced during the image generation process~\cite{sun}.
For example, the fingerprints generated during the upsampling estimation of the generator can be successfully detected in the frequency domain~\cite{frank,watch_cvpr20,unmasking,chen2}.
Unfortunately, the frequency-level fingerprints differ between generative models and object categories, causing GAN detectors to be data-dependent. 
As a result, when the GAN detectors are tested with the generated images of the unseen GAN models or object categories, they experience a performance reduction~\cite{easy_to_detect}.
Although \cite{cozzolino,tgd} tried to overcome these limitations by using transfer learning with a little amount of data, it is unreasonable to expect to be able to get training data for the unknown GAN models employed by the abusers~\cite{huh}.  
For the generalization of GAN detectors, it is far more efficient to use unsupervised or self-supervised training.

Therefore, we suggest a novel self-supervised framework composed of the artificial fingerprint generator and the GAN detector.
The artificial fingerprint generator is trained only with the real images to reconstruct the diverse artificial fingerprints of the generated images by GANs.
Based on the well-known analysis that the upconvolution layers produce the artifacts~\cite{watch_cvpr20,upconv_dzanic,upconv_chand}, we build our artificial fingerprint generator to be composed of multiple autoencoders effectively reproducing the diverse artificial fingerprints of GAN models.
To further consider the GAN models where the upconvolution layers are removed~\cite{biggan,stargan2}, we also utilize the additional autoencoder designed to reconstruct the input image without any upsampling operation.
Then, the GAN detector is trained to distinguish the real images and the reconstructed images from the artificial fingerprint generator, so we can ignore the fake images of the training datasets in the overall training processes.
Unlike unsupervised GAN networks~\cite{gan}, which generate images of varying quality based on learning the data distributions of real images, autoencoders can reconstruct high-quality images that are nearly identical to the original real images while effectively avoiding data dependency due to a specific GAN dataset.
Our method is tested with various real-world scenarios to validate the state-of-the-art generalization ability of our model to detect the unseen GAN models and object categories.
In addition, our additional experiments reveal that when extra target fake images are included in the GAN detector's training, our approach shows reasonable improvements compared to the previous studies.

We can summarize our contributions as follows:
\begin{itemize} \label{contribution}
  \item Unlike the previous GAN detectors dependent on the generated images by GAN models, our model utilizes the self-supervised training method to obtain generalized detection ability and avoid data dependency.
  \item We build the artificial fingerprint generator that replicates the generation of artifacts from the various GAN models independent of their upconvolution layer composition, allowing the GAN detector to be trained robustly across different GAN models and categories.
  \item Our method shows robust performance compared to the previous GAN detectors even without training the fake images generated by GAN models.
\end{itemize}

%% file: 2_related_work.tex
\section{Related Work}
Image-based detection, frequency-based detection, and transfer learning for detection are the three categories of earlier literature on the detection of created pictures.

\begin{figure*}[]
\centering
\includegraphics[width=0.6\linewidth]{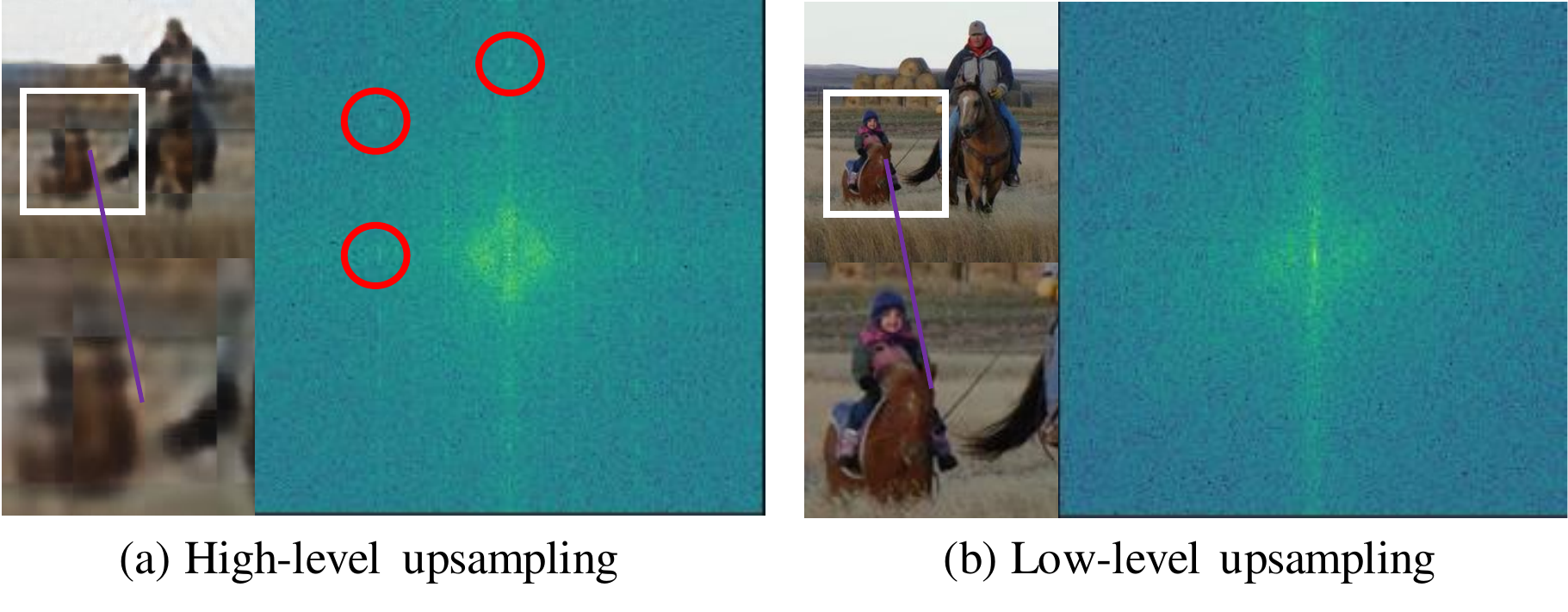}
\caption{\textbf{The reconstructed images of the autoencoders and their 2D spectra.}
(a) The high-level upsampling image of the white box and its 2D spectrum with the artificial fingerprints in the red circles, and (b) the low-level upsampling image of the white box and its 2D spectrum. The frequency-level fingerprints appear frequently in high-level upsampling, while the box size of pixel-level checkerboard artifacts is enlarged.}
\label{fig:image_up_sample}
\vspace{-1.5em}
\end{figure*}

\subsection{Image-based Detection}
To begin, pixel-level characteristics of GAN-based generated images can be employed as identifiers to differentiate between genuine and generated images.
The identifiers are referred to as 'artifacts,' and they are formed as a result of the generator's operations in GANs~\cite{easy_to_detect,watch_cvpr20,chen2}. 
According to some prior research, discrepancies in blocking artifacts formed during JPEG compression~\cite{ye, tralic} or demosaicing artifacts created by a color filter array~\cite{dirik, ferrara} should be investigated.
Blob-shaped abnormalities, which can be seen in the produced photos of \cite{stylegan}, are one of the most well-known pixel-level artifacts.
The blob-shaped artifacts exist in apparent shapes in the generator's intermediate feature maps, even though they become extremely subtle in the final image~\cite{stylegan2}.
Other image-based detection methods include an adaptable autoencoder-based neural network architecture for new target domains~\cite{cozzolino} and cross-model manipulation detection using post-processing techniques like JPEG and blur~\cite{adobe}.

\subsection{Frequency-based Detection}
For improved performance, a number of GAN detectors focus on the distinctive patterns in the frequency spectra.
\cite{kirchner} proposes combining the variance of the prediction residue with artifacts in the spatial, frequency domain, whereas \cite{huang} proposes using the Fast Fourier Transform~\cite{fft} and singular value decomposition to distinguish image manipulations like JPEG compression, Gaussian noise, and blur.
In addition, \cite{marra} uses artificial fingerprints to detect forged regions using frequency-based, GAN-specific detection, whereas \cite{bappy} suggests a manipulation localization utilizing frequency domain correlation and geographical maps.
Recently, \cite{frank} considers utilizing the Discrete Cosine Transform~\cite{dct} to analyze GAN-related artifacts in frequency space, while \cite{zhang} provides a detection approach based on the artifacts caused by the up-sampler of GANs.
In addition, \cite{watch_cvpr20, unmasking} suggests using Azimuthal integration to leverage spectral aberrations for detecting inauthentic photos.
However, the frequency-level fingerprints are difficult to generalize since the design of the generators can vary their appearance.

\subsection{Transfer Learning for GAN Detector}
It is impossible to update the model's training in a supervised manner on a daily basis since new manipulation methods arise on a daily basis~\cite{aneja}. Transfer learning may be used to improve the generalization of GAN detectors to prevent this problem, according to \cite{cozzolino,tgd,aneja}.
Transfer learning is the process of using a model that has already been trained for one job to do another task with less data.
Transfer learning can enable the GAN detector to successfully adapt to new domains utilizing a few samples of generated images during training, as proposed by \cite{cozzolino} in Forensic-Transfer.
To improve detection performance, \cite{tgd} recently suggested a transferable framework for GAN detectors based on self-training, which successfully recognizes generated images with a little amount of data.

%% file: 3_method.tex
\section{Self-supervised GAN Detector} \label{sec:framework}
As illustrated in Fig.~\ref{fig:framework}, our framework comprises of two modules: the artificial fingerprint generator and the GAN detector.
The artificial fingerprint generator is trained to reconstruct the input images in the unsupervised scheme, while the GAN detector is trained to distinguish the origin real images from the images reconstructed by the artificial fingerprint generator.
The artificial fingerprint generator is trained first, and the GAN detector is trained by using the images generated by the trained generator.

\subsection{Artificial Fingerprint Generator} 
As shown in Fig.~\ref{fig:image_up_sample}, since the appearance of the frequency-level fingerprints can vary by the number of upsampling operations in the generator network of GAN models, our proposed framework should be able to distinguish the various appearances for generalized detection.
Thus, we design the generator using multiple autoencoders with a diverse number of upsampling operations, which is named as the \textit{`artificial fingerprint generator.'}
For the deconvolution-based networks, using the autoencoders with the entire scales can enlarge the size of the framework and also slow down the training process; thus, we design an effective framework by mixing up the two artificial fingerprints generated by the large and small number of convolution layers, respectively. 
Also, to manage the interpolation-based GAN models~\cite{biggan,stargan2}, we employ an additional autoencoder that reconstructs images without any upsampling operation.
As a consequence, the artificial fingerprint generator consists of three autoencoders, which include the autoencoders containing the high and low levels of deconvolution layers respectively, and an autoencoder without any upsampling operation.

The first autoencoder is the high-level upsampling autoencoder ($G_{high}$), which applies the downsampling process of 1/64 to the input images and then applies the upsampling process to reconstruct the original inputs. 
The second autoencoder is the low-level upsampling autoencoder ($G_{low}$), which applies the downsampling process of 1/2 and then applies the upsampling process.
The upsampling blocks of the first and second autoencoders are built by a deconvolution layer and the activation function of ReLU~\cite{relu}. 
Since we design every convolution layer of the autoencoder with the stride 2 for downsampling, $t$ times of downsampling for $2^t=d$ must be proceeded to reach the size of $1/d$. The following deconvolution layers are also built with stride 2, so $t$ times of upsampling must be proceeded for the reconstruction.
Thus, $G_{high}$ contains an encoder of 6 convolution layers and a decoder of 6 transposed convolution layers~\cite{transconv}, while only one convolution layer and one transposed convolution layer are used in $G_{low}$.

The third autoencoder is the non-upsampling autoencoder ($G_{non}$), which is to detect the images without applying the deconvolution process. 
To avoid applying the deconvolution process, only the convolution layer is used, and the strides are set to 1 to avoid upsampling operations.
$G_{non}$ is composed of two convolution layers respectively for the encoder and the decoder.

The training set for the three autoencoders consists of the real images only. The training loss is a reconstruction loss using the mean squared error as:
\begin{equation}\label{eq:rec_loss}
        \mathcal{L}(X, G)=\mathbb{E}_{x\sim X}[||x-G(x)||_{2}],
\end{equation}
where $X$ is the training set containing the samples $x$ and $G$ is one of the autoencoders including $G_{high}$, $G_{low}$, and $G_{non}$.
The three autoencoders are individually trained using the same datasets, but they generate different types of artificial fingerprints due to various levels of upsampling operations.

\subsection{GAN Detector}
After finishing the training of the artificial fingerprint generator, we train the GAN detector to distinguish the generated images from the real images.
The frequency-level analysis is effective in observing the artificial fingerprints. 
Thus, we also utilize the 2D spectrum.
% by Fast Fourier Transform (FFT)~\cite{fft}. The generated image $\hat{x}\in\hat{X}=\{G_{high}(x), G_{low}(x), G_{non}(x)\vert \forall x \in X\}$ of the artificial fingerprint generator is transformed into a 2D spectrum by FFT. 

For the fair comparison with the recent models~\cite{frank,watch_cvpr20,adobe}, we design our GAN detector based on ResNet-50~\cite{resnet}.
To train the GAN detector, we reconstruct all real images of the training dataset, which is considered as the generated images.
Since the three autoencoders reconstruct three images from a real image, the training of GAN detector can suffer from the unbalancing problem.
Thus, when we compose the mini-batch, the sampling probability is adjusted to extract the real images three times more frequently than the reconstructed images.

In addition to the raw images, we utilize the mixup algorithm~\cite{zhang2018mixup} to consider the mixed artificial fingerprints by integrating all three types of the artificial fingerprints with the real images.
When we define two samples randomly selected from a mini-batch ($\mathbf{S}=\{X, \hat{X}\}$) by $\mathbf{s}_i$ and $\mathbf{s}_j$ and their corresponding one-hot labels are $\mathbf{y}_i$ and $\mathbf{y}_j$, respectively, the mixed sample $\tilde{\mathbf{s}}_{(i,j)}$ and its corresponding label $\tilde{\mathbf{y}}_{(i,j)}$ are obtained by:
\begin{equation}\label{eq:label}
%\begin{aligned}
    \tilde{\mathbf{s}}_{(i,j)} = \lambda \mathbf{s}_i + (1 -\lambda) \mathbf{s}_j, \hspace{1cm}
    \tilde{\mathbf{y}}_{(i,j)} = \lambda \mathbf{y}_i + (1 -\lambda) \mathbf{y}_j,
%\end{aligned}
\end{equation}
where $\lambda\in[0,1]$ is a mixing scale randomly chosen by a Beta distribution.

Then, we replace every sample of the mini-batch with the mixed samples by iterating the mixup algorithm, which is defined by $\tilde{\mathbf{S}}$. The labels corresponding to the samples of $\tilde{\mathbf{S}}$ is defined as $\tilde{\mathbf{Y}}$.
Then, the GAN detector~($C$) is trained by a conventional softmax cross-entropy loss as:
\begin{equation}\label{eq:updateD}
\begin{aligned}
    \mathcal{L}_{C}(\tilde{\mathbf{S}}) = \mathbb{E}_{(\tilde{\mathbf{s}},\tilde{\mathbf{y}}) \sim (\tilde{\mathbf{S}},\tilde{\mathbf{Y}})}[\tilde{\mathbf{y}}^T \log(C(\tilde{\mathbf{s}}))].
\end{aligned}
\end{equation}

\begin{table}[]
    \begin{center}
    \caption{\textbf{Cross-category performance.} }
    \resizebox{0.7\linewidth}{!}{%
        % \centering
        \begin{tabular}{cccc}

        \hline
        & & Acc. & A.P. \\ \hline
        & Wang~\cite{adobe} & 50.4 & 63.8\\
        Supervised & Frank~\cite{frank} & 78.9 & 77.9 \\
        & Durall~\cite{watch_cvpr20} & 85.1 & 79.5 \\\hline
        Unsupervised & Ours & \textbf{92.0} & \textbf{97.7} \\\hline
        \end{tabular}
    }
    \vspace{-5mm}
    \label{tab:cross_objs}
    \end{center}
\end{table}

\begin{table*}[h!]
\centering
\scriptsize
\caption{\textbf{Zero-shot cross-model performance.} }
\resizebox{1.00\linewidth}{!}{%
\begin{tabular}{lccccccccccccccc|cc}
\hline
\multicolumn{1}{c}{\multirow{2}{*}{Model}} & \multicolumn{1}{c}{\multirow{2}{*}{\begin{tabular}[c]{@{}c@{}}number of training class\\ (real/fake)\end{tabular}}} & \multicolumn{2}{c}{StyleGAN} & \multicolumn{2}{c}{StyleGAN2} & \multicolumn{2}{c}{BigGAN} & \multicolumn{2}{c}{CycleGAN} & \multicolumn{2}{c}{StarGAN} & \multicolumn{2}{c}{GauGAN}& \multicolumn{2}{c}{Deepfake} & \multicolumn{2}{|c}{Mean}\\ \cline{3-18} 
& & Acc. & A.P. & Acc. & A.P. & Acc. & A.P. & Acc. & A.P. & Acc. & A.P.& Acc. & A.P. & Acc. & A.P. & Acc. & A.P. \\ \hline
Wang\cite{adam} & (1,1) & 50.4 & 79.3 & 68.2 & 94.7 & 50.2 & 61.3 & 50 & 52.9 & 50 & 58.2 & 50.3 & 67.6 & 50.1 & 51.5 & 52.7  & 66.5 \\
Frank~\cite{frank} & (1,1) & 68.5 & 80.7 & 60.8 & 77.3 & 72.1 & 63 & 57.6 & 56.6 & 80.1 & 76.3 & 74 & 95.5 & 54.4 & 59.4 & 66.8  & 72.7 \\
Durall~\cite{watch_cvpr20} & (1,1) & 64.1 & 58.6 & 69.3 & 62.9 & 55.4 & 52.9 & 69.6 & 62.8 & 95.4 & 91.5 & 57.5 & 54 & 53.6 & 52 & 66.4  & 62.1 \\
\hline
Wang\cite{adam} & (2,2) & 52.8 & 82.8 & 75.7 & 96.6 & 51.6 & 70.5 & 58.6 & 81.5 & 51.2 & 74.3 & 53.6 & 86.6 & 50.6 & 51.5 & 56.3  & 77.7 \\
Frank~\cite{frank} & (2,2) & 70.8 & 83.8 & 61.2 & 75.6 & 74.9 & 76.2 & 74.8 & 76.8 & 91.7 & 97.5 & 89.2 & 98.4 & 52.8 & 53 & 73.6  & 80.2 \\
Durall~\cite{watch_cvpr20} & (2,2) & 63.5 & 58.1 & 68.7 & 62.4 & 56.4 & 53.5 & 63.5 & 58.2 & 89.8 & 83.1 & 56.5 & 53.5 & 53.3 & 51.7 & 64.5 & 60.1 \\
\hline
Wang\cite{adam} & (4,4) & 63.8 & 91.4 & 76.4 & 97.5 & 52.9 & 73.3 & 72.7 & 88.6 & 63.8 & 90.8 & 63.9 & 92.2 & 51.7 & 62.3 & 63.6  & 85.2 \\
Frank~\cite{frank} & (4,4) & 72.2 & 82.1 & 64.2 & 80.1 & 68.9 & 82.4 & 53.7 & 66.2 & 89.1 & 99.2 & 65.3 & 90.3 & 51.1 & 49.6 & 66.4  & 78.6 \\
Durall~\cite{watch_cvpr20} & (4,4) & 63.9 & 58.4 & 69 & 62.7 & 58.5 & 54.7 & 69.6 & 63.1 & 99 & 98.1 & 57 & 53.8 & 50.4 & 50.1 & 66.8  & 63.0 \\
\hline
Wang\cite{adam} & (20,20) & 71.4 & 96.3 & 67.5 & 93.4 & 60.9 & 83.3 & 83.8 & 94.3 & 84.6 & 93.6 & 79.3 & 98.1 & 51.1 & 66.3 & 71.2  & \textbf{89.3} \\
Frank~\cite{frank} & (20,20) & 81.8 & 91.7 & 71.4 & 93.0 & 76.0 & 87.8 & 62.8 & 77.3 & 96.9 & 99.4 & 73.9 & 93.1 &  48.6 & 48.3 & 73.1 & 84.4 \\
Durall~\cite{watch_cvpr20} & (20,20) & 64.7 & 59.0 & 69.2 & 62.9 & 59.4 & 55.3 & 66.9 & 60.9 & 98.5 & 97.1 & 57.2 & 53.9 & 52.2 & 51.3 & 66.9 & 62.9 \\
\hline
Our & (1,0) & 71.5 & 79.1 & 70.0 & 79.3 & 77.3 & 90.0 & 87.5 & 86.3 & 99.8 & 100. & 70.9 & 96.9 & 69.2 & 74.0 & \textbf{78.0} & 86.5 
 \\
\hline
\end{tabular}
}
\label{tab:cross_gan_supple}
\end{table*}

\begin{figure*}[t!]
\centering
\includegraphics[width=0.8\linewidth]{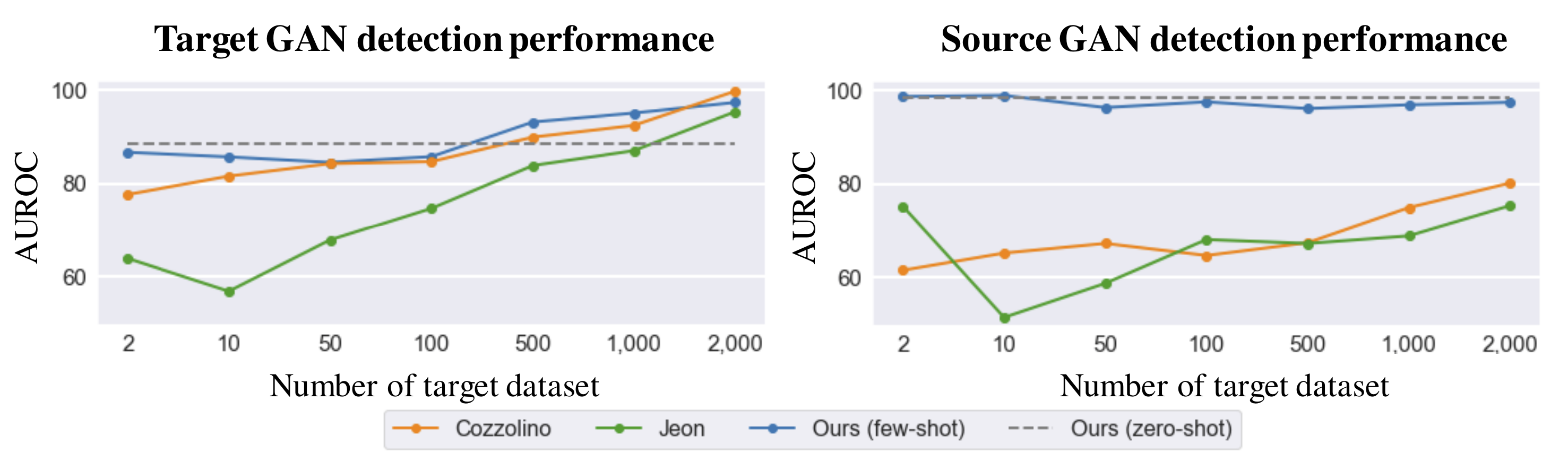}
\caption{\textbf{Performance comparison on transfer learning.} The performance comparison of GAN detectors based on transfer learning is shown according to a gradual increase of training data for the target and the source, respectively.
The target GAN performance is shown on the left, and the source GAN performance is shown on the right.
}
\label{fig:transfer_test} 
\end{figure*}

\begin{table*}[]
\centering
\scriptsize
\caption{\textbf{Few-shot learning performance. } }
\resizebox{1.00\linewidth}{!}{%
\begin{tabular}{lccccccccccccccc|cc}
\hline
\multicolumn{1}{c}{\multirow{3}{*}{Model}} & \multicolumn{1}{c}{\multirow{3}{*}{Shot}}  & \multicolumn{16}{c}{Test Models} \\ \cline{3-18} 
&  & \multicolumn{2}{c}{StyleGAN} & \multicolumn{2}{c}{StyleGAN2} & \multicolumn{2}{c}{BigGAN} & \multicolumn{2}{c}{CycleGAN} & \multicolumn{2}{c}{StarGAN} & \multicolumn{2}{c}{GauGAN}& \multicolumn{2}{c}{Deepfake} & \multicolumn{2}{c}{Mean} \\ \cline{3-18} 
 &  & SRC & TRG &   SRC & TRG &   SRC & TRG &   SRC & TRG&   SRC & TRG &   SRC & TRG &   SRC & TRG &   SRC & TRG \\ \hline
Cozzolino~\cite{cozzolino} & {1,000} & 96.8 & 99.2 & 71.3 & 76.1 & 76.1 & 94.0 & 68.5 & 99.2 & 62.6 & 100. & 84.0 & 99.9 & 65.3 & 78.2 & 74.9 & 92.3 \\
Cozzolino~\cite{cozzolino} & {2,000} & 98.8 & 99.7 & 96.0 & 99.9 & 82.6 & 99.8 & 65.2 & 99.9 & 62.4 & 100. & 76.7 & 99.9 & 79.5 & 98.4 & 80.1 & \textbf{99.6} \\
Jeon~\cite{tgd}& {1,000} & 67.7 & 94.0 & 60.0 & 89.5 & 59.7 & 66.2 & 71.0 & 82.5 & 80.2 & 99.3 & 94.6 & 97.2 & 49.3 & 80.2 & 68.9 & 86.9 \\
Jeon~\cite{tgd} & 2,000 & 64.3 & 96.8 & 69.3 & 99.4 & 64.2 & 85.8 & 98.3 & 98.2 & 85.4 & 99.8 & 96.7 & 98.4 & 49.5 & 88.6 & 75.3 & 95.2 \\
\hdashline
 Ours & {0} & 98.3 & 76.5 & 98.3 & 80.3 & 98.3 & 91.9 & 98.3 & 89.1 & 98.3 & 100. & 98.3 & 97.6 & 98.3 & 74.7 & \textbf{98.3} & 88.6 \\
 Ours & {1,000} & 97.9 & 94.3 & 97.3 & 95.0 & 96.7 & 98.6 & 95.0 & 96.8 & 98.1 & 100. & 95.5 & 99.8 & 97.6 & 80.6 & 96.9 & 95.0 \\
 Ours & {2,000} & 98.0 & 97.3 & 98.6 & 98.4 & 96.6 & 99.1 & 95.1 & 99.4 & 97.9 & 100. & 97.3 & 99.8 & 98.3 & 86.9 & 97.4 & 97.3 \\
\hline
\end{tabular}
}
\label{tab:transfer}
\end{table*}

\begin{table*}[]
%\vspace{-0.5em}
\centering
\scriptsize
\caption{\textbf{Ablation study for self-supervised GAN detector.} }
\resizebox{1.00\linewidth}{!}{%
\begin{tabular}{lcccccccccccccccc|cc}
\hline
\multicolumn{1}{c}{\multirow{3}{*}{Model}}  & \multicolumn{14}{c}{Test Models} \\ \cline{2-19} 
& \multicolumn{2}{c}{{ProGAN}} & \multicolumn{2}{c}{{StyleGAN}} & \multicolumn{2}{c}{{StyleGAN2}} & \multicolumn{2}{c}{{BigGAN}} & \multicolumn{2}{c}{{CycleGAN}} & \multicolumn{2}{c}{{StarGAN}} & \multicolumn{2}{c}{{GauGAN}}& \multicolumn{2}{c}{{Deepfake}} & \multicolumn{2}{|c}{{Mean}}\\ \cline{2-19} 
 & Acc. & A.P. & Acc. & A.P. & Acc. & A.P. & Acc. & A.P. & Acc. & A.P.& Acc. & A.P. & Acc. & A.P. & Acc. & A.P. & Acc. & A.P. \\ \hline
Ours & 94.1 & 98.3 & 71.5 & 79.1 & 70.0 & 79.3 & 77.3 & 90.0 & 57.5 & 86.3 & 99.8 & 100. & 70.9 & 96.9 & 69.2 & 74.0 & \textbf{76.3} & \textbf{88.0} \\
w/o~{$G_{high}$} & 92.4 & 98.2 & 67.2 & 77.8 & 72.6 & 80.0 & 71.5 & 91.1 & 53.9 & 88.9 & 99.7 & 100. & 63.5 & 84.9 & 55.3 & 58.3 & 72.0 & 84.9 \\
w/o~{$G_{low}$} & 90.1 & 97.9 & 68.1 & 81.6 & 61.5 & 70.2 & 66.2 & 77.3 & 52.0 & 47.6 & 98.7 & 99.9 & 67.1 & 83.3 & 52.9 & 53.7 & 69.6  & 76.4 \\
w/o~{$G_{non}$} & 54.8 & 88.6 & 55.0 & 75.1 & 59.3 & 85.3 & 85.9 & 92.0 & 80.0 & 94.0 & 100. & 100. & 85.0 & 95.7 & 50.1 & 56.8 & 71.3  & 85.9 \\
w/o~{Freq.} & 91.1 & 98.0 & 71.3 & 80.3 & 67.8 & 77.8 & 73.0 & 77.1 & 61.4 & 90.5 & 99.4 & 100. & 71.3 & 94.8 & 65.4 & 70.6 & 75.1  & 86.1 \\
w/o~{Mixup} & 92.1 & 97.6 & 67.7 & 76.8 & 66.2 & 71.4 & 70.9 & 78.8 & 60.2 & 87.4 & 93.7 & 100. & 69.4 & 93.4 & 56.0 & 59.6 & 72.0  & 83.1 \\

\hline
\end{tabular}
}
\label{tab:ablation}
\end{table*}

%% file: 4_result.tex
\section{Experimental Results}
\subsection{Dataset} \label{dataset}
Through experiments, we compare the performance of each network by employing the same data. Since the training settings have a strong impact on the analysis of the GAN detector, we utilize the real horse images of LSUN~\cite{lsun}, which is the dataset used for the training of ProGAN~\cite{progan}.
In contrast, the comparing models use the generated images of horse, car, cat, and airplane categories of ProGAN~\cite{progan}, as well as the real images of LSUN~\cite{lsun}.
We utilize the 20 different object categories of ProGAN~\cite{progan} to observe the changes in the performance of the GAN detectors based on the change in categories of the same model.

Also, to evaluate the level of data dependency due to a specific GAN dataset, we utilize the most well-known unconditional GAN models with high resolutions, such as ProGAN~\cite{progan}, StyleGAN~\cite{stylegan}, and StyleGAN2~\cite{stylegan2}. 
Then, to evaluate the performance of the GAN detectors on generating new images from the latent space, we employ the conditional GAN model named BigGAN~\cite{biggan}, which is known to realistically reconstruct the data of the most categories, and the image-to-image translation models, such as CycleGAN~\cite{cyclegan}, StarGAN~\cite{stargan}, and GauGAN~\cite{gaugan}.
Lastly, we conduct experiments to test the models whether they can detect not only the entire synthesis of face images but also partially generated images using FaceSwap~\cite{faceforensics++}. We utilize various GANs with human faces and various objects, and the real images used to train the GANs, including CelebA-HQ~\cite{celebahq}, CelebA~\cite{celeba}, COCO~\cite{coco}, LSUN~\cite{lsun}, and ImageNet~\cite{imagenet}, which are all publicly available through~\cite{adobe} and~\cite{tgd}.

\subsection{Evaluation Metrics}
For performance comparison, we employ the accuracy and average precision~\cite{VOC2010}, which are the metrics commonly used in this field of study. To test the transfer-learning performance, we use the Area Under the Receiver Operating Characteristic (AUROC) as used in~\cite{tgd,cozzolino}. 
To compare the generalization performance, we follow the suggestion of Wang~\etal~\cite{adobe} to use JPEG compression, which is known as the most effective method to test the generalization performance. Also, for the frequency-level analysis, we compare with Frank~\etal~\cite{frank} and Durall~\etal~\cite{unmasking}, both of which study the frequency-based detection. Lastly, for zero-shot and domain transfer performance, we compare with Jeon~\etal~\cite{tgd} and Cozzolino~\etal~\cite{cozzolino}.

\subsection{Implementation Details} \label{imple_detail}
We resize the input images into 256$\times$256, and the reconstructed images with the same size. For training, we use a single NVIDIA RTX 8000 with a batch size of 16 and 20 epochs for the GAN detector, 80 epochs for the artificial fingerprint generator. Both of the artificial fingerprint generator and GAN detector networks are trained by Adam optimizer~\cite{adam} with the learning rate of 0.0001, which are the conventional hyperparameters of the previous GAN detectors~\cite{adobe, bihpf}.

\subsection{Generalization Performance of the GAN Detector} 
We train the GAN detectors with only one object category (\textit{horse}) of real images and test with the entire object categories and GAN models to evaluate the generalization performance.
We perform two different evaluation to show the generality of our method to the unseen object categories and the unseen GAN models, respectively.
We compare with the previous studies used for the comparison: Wang~\etal~\cite{adobe}, Frank~\etal~\cite{frank}, and Durall~\etal~\cite{watch_cvpr20}.

At first, the generality to the unseen object categories is evaluated by using the training data of \textit{horse} category.
While our method utilizes only the real images of \textit{horse} category, the compared models additionally use the fake images generated from ProGAN since they should be trained in a supervision scheme.
As shown in Table~\ref{tab:cross_objs}, when we evaluate the trained models for the remaining categories of ProGAN dataset, our method outperforms the compared methods without considering any fake images generated from ProGAN.
This result shows that our method can generate the general artifacts to cover even the unseen object categories by using only the real images.

Then, to show the generality of our method for the unseen GAN models, we evaluate the zero-shot cross-model performance.
The proposed model is trained by the real images of \textit{horse} category as like the experiments of unseen object categories.
In contrast to the proposed algorithm, since the previous studies cannot be trained without the fake images, they need to be trained by using the fake data of the training dataset.
For the compared methods, we perform repeated experiments with the various numbers of the training categories of ProGAN~\cite{progan}. 
As presented in Table~\ref{tab:cross_gan_supple}, even with the highly limited setting, our GAN detector achieves the highest accuracy and average precision for the unseen GAN models, even without using any generated images of GAN models.
This result validates the robustness of our method to the unseen GAN models by integrating the three autoencoders to generate the artificial fingerprints.

\subsection{Transfer GAN Domain Performance}
In this section, we compare the zero-shot performance of our model and other models employing transfer learning, which is also known as few-shot learning. 
The comparing models follow the standards of \cite{tgd,cozzolino} to utilize the 1,000 to 2,000 generated images of a specific GAN model for transfer learning, and ProGAN~\cite{progan} as the source model. To apply transfer learning to our model, after finishing the baseline training of our GAN detector, we fine-tune the model by 1 epoch with the mixed datasets composed of the additional target dataset and the reconstructed data.
The proportion of the target GAN domain in the entire batch is set to 15\%.

Fig.~\ref{fig:transfer_test} shows the performance comparison of the models trained with 2 to 2,000 real and generated images to observe whether the models maintain the performance in the target and source GAN as the number of training images increase.
As shown in Table~\ref{tab:transfer}, our model not only achieves competitive performance compared to other models employing transfer learning in the target domain (TRG) but also outperforms others in the source domain (SRC) by additionally using the reconstructed images of the artificial fingerprint generator. 
Overall, our model maintains robust performance in the source GAN while achieving increased performance in the target GAN. In the experiment of few-shot learning with the training images under 1,000, our model achieves the most robust performance, and when the training images are expanded to 2,000, our model achieves 97.4\% in AUROC.

\subsection{Ablation Study}
Table~\ref{tab:ablation} shows the individual effect of each component of our proposed method. In the second row, `w/o $G_{high}$' indicates the results of downsampling to 1/64 without applying the artificial fingerprint generator for upsampling. The third and fourth columns also show the results without applying $G_{low}$, $G_{non}$. 
Also, it shows the results of using the pixel-level analysis instead of the frequency-level, and the results without the mix-up algorithm.
By excluding the data without the downsampling process, the performances of ProGAN, StyleGAN, and StyleGAN2 dramatically decline, while those of BigGAN, CycleGAN, and GauGAN increase. 
This indicates the artificial fingerprints generated by the transposed convolution are relatively few in the images of BigGAN, CycleGAN, and GauGAN. 
Thus, in order to generally detect images generated by various GAN models, it is important to utilize the images reconstructed from the autoencoder without the upsampling process.
Also, the mixup algorithm enables the artificial fingerprint generator to generate various sizes of artificial fingerprints, which boosts up the training effect.
Interestingly, the achievement from the frequency-level transformation seems minor compared to the other components, allowing us to claim that the artificial fingerprints can be detected effectively even with the pixel-level images.

%% file: 5_conclusion.tex
\section{Conclusion} \label{conclusion}
We propose a self-supervised GAN detector with robust generalization ability to detect the generated images even upon the unseen setting during training. Through a comprehensive analysis of the frequency-level fingerprints found in the previous studies, we build the integrated framework composed of the artificial fingerprint generator and GAN detector. By reconstructing the artificial fingerprints in the frequency domain and utilizing them as the training data, our proposed method achieves state-of-the-art performance even without using any generated images by GANs. Especially, our method shows impressive performance when tested with unseen categories and GAN models, which validates the robust generalization of our GAN detector. Based on the various frequency-level fingerprints of the generated images, we plan to analyze the missing characteristics of the fingerprints to further extend our framework.

\section*{Acknowledgments}
This work was partly supported by Samsung SDS, Institute of Information \& communications Technology Planning \& Evaluation (IITP) grant funded by the Korea government (MSIT) (2021-0-01341, Artificial Intelligence Graduate School Program(Chung-Ang University)), and the Chung-Ang University Research Grants in 2020.